\documentclass[11pt,a4paper]{article}
\usepackage[final]{acl2021}
\usepackage{times}
\usepackage{latexsym}

\usepackage{microtype}
\usepackage{hyperref}
\usepackage{url}
\usepackage{tabularx}
\usepackage{times}
\usepackage{latexsym}
\usepackage{booktabs}
\usepackage{url}
\usepackage{graphicx}
\usepackage{xcolor}
\usepackage{multirow}

\usepackage{microtype}
\usepackage{arydshln}
\usepackage{amsmath}
\usepackage{amssymb}
\usepackage{enumitem}

\title{Cooperative Self-training of Machine Reading Comprehension}

\author{Hongyin Luo$^1$ $    $ Shang-Wen Li$^{2}$ $  $ Mingye Gao$^{3}$ $  $ Seunghak Yu$^2$\thanks{ * Work is not related to the employment of the second and fourth authors at Meta and Amazon} $    $ James Glass$^1$ \\
  $^{1}$MIT CSAIL, $^{2}$Amazon AI  $^{3}$MIT MTL\\
  {\tt \{hyluo,mingye,glass\}@mit.edu,\{shangwel,yuseungh\}@amazon.com}\\
}

\date{}

\begin{document}
\maketitle

\begin{abstract}
Pretrained language models have significantly improved the performance of downstream language understanding tasks, including extractive question answering, by providing high-quality contextualized word embeddings. However, training question answering models still requires large amounts of annotated data for specific domains. In this work, we propose a cooperative self-training framework, RGX, for automatically generating more non-trivial question-answer pairs to improve model performance. RGX is built upon a masked answer extraction task with an interactive learning environment containing an answer entity \textbf{R}ecognizer, a question \textbf{G}enerator, and an answer e\textbf{X}tractor. Given a passage with a masked entity, the generator generates a question around the entity, and the extractor is trained to extract the masked entity with the generated question and raw texts. The framework allows the training of question generation and answering models on any text corpora without annotation.
We further leverage a self-training technique to improve the performance of both question generation and answer extraction models.
Experiment results show that RGX outperforms the state-of-the-art (SOTA) pretrained language models and transfer learning approaches on standard question-answering benchmarks, and yields the new SOTA performance under given model size and transfer learning settings.

\end{abstract}

\section{Introduction}
\label{sec:intro}
\begin{figure}[t]
\centering
\includegraphics[width=\columnwidth]{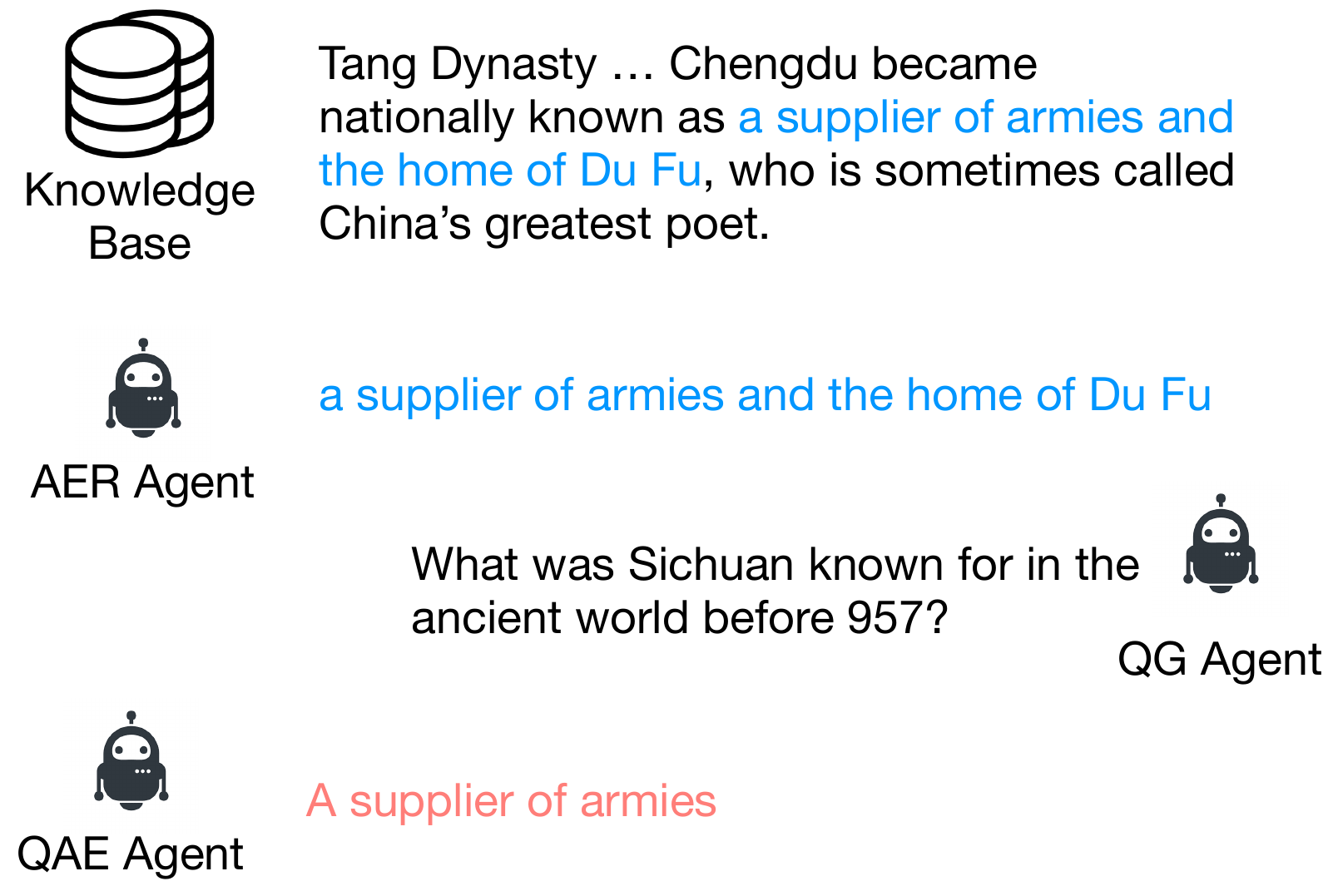}
\caption{The pipeline of semi-supervised question answering (machine reading comprehension) by RGX. AER (answer entity \textbf{R}ecognition) agent recognizes answer entity from a given passage; QG (question \textbf{G}eneration) generates a question based on the passage and entity; QAE (question-answering e\textbf{X}tractor) extracts answer from the question and passage.}
\label{fig:example}
\end{figure}

Recent studies have shown that language model pretraining provides high-quality text representations and significantly improves neural networks' performance on a variety of natural language processing (NLP) tasks \cite{peters2018deep}. Based on the popular Transformer architecture \cite{vaswani2017attention}, various language models have been proposed \cite{devlin2018bert,liu2019roberta,clark2019electra}. These models are pretrained to predict a masked word in a given context from large corpora, and generate a contextual representation that encodes semantic and syntactic information. After finetuning, these representations significantly improve performance on downstream NLP tasks. Although masked language modeling is a powerful self-supervised learning technique, annotation on large-scaled data is still necessary for finetuning on difficult downstream tasks, including extractive question answering (QA)\footnote{Also referred to as machine reading comprehension. The two terms are used interchangeably in this paper.} where a large number of labeled question-answer pairs are required as a training corpora.


Previous studies showed that the QA models can be improved by training on synthetic question-answer pairs, namely self-training \cite{sachan2018self,puri2020training,shakeri2020end,bartolo2021improving}. The core step of these work is pretraining a question-answer pair synthesis model on a seed corpus, and apply the generator on target domains to obtain synthetic training data. The QA model learns domain knowledge after finetuning on the synthetic data, and thus the domain adaptation is improved. However, the gap between the pretraining (i.e., seed) and the target corpus still exists, in terms of domain knowledge, question difficulty, and language style. The gap affects the quality of the synthetic training data.

We thus propose a framework that allows cooperative self-training for both QA pair synthesis and question answering to better adapt the synthesis models to the target domain and improve the learning of the QA models. In the framework, we construct a cooperative environment where a question generator and an answer extractor work together  to solve a masked entity prediction problem. We first leverage an entity recognizer to mask out an entity in a provided passage. The question generator then outputs a question based on the masked passage. With the generated question and the original, unmasked passage, we train the answer extractor to select the correct answer spans, which are the masked entity. The extractor is also the final model used for extractive QA. To extract the spans accurately, the generator has to provide a good question, and the extractor should select the most likely tokens.
We apply an expectation-maximization algorithm to select high-quality QA pairs and update both question generation and answer extraction models to improve the quality of synthetic data and the accuracy of the self-trained QA model based on synthetic QA pairs.
We call our algorithm RGX since it incorporates an answer entity \textbf{R}ecognizer, a question \textbf{G}enerator, and a question-answering e\textbf{X}tractor. The RGX pipeline is illustrated in Figure \ref{fig:example}.


With RGX, we can train a QA model for any unlabeled target domain given the corresponding text corpora and a labeled QA corpus in a seed domain (either the same or different from the target). 
By training QA models on synthetic QA data generated by RGX and evaluating the trained model on human-labeled evaluation data, we show that RGX outperforms SOTA approaches in QA benchmark datasets when domain specific human labels are not available during finetuning. In this work, we make the following contributions:
\begin{enumerate} \setlength{\itemsep}{0pt} \setlength{\parsep}{0pt}
\item We propose a cooperative self-training framework, RGX, which contains an answer entity recognition, question generation, and answer span extraction to automatically generate non-trivial QA pairs on unlabeled corpora.
\item We design a expectation-maximization (EM) synthetic QA selection that identifies difficult but answerable questions without supervision to incrementally train the QA model with challenging examples, and an answer entity recognition (AER) based maximum mutual information (MMI) inference method for question answering.
\item Experiments show that our method significantly outperforms SOTA pretrained QA models and self-training QA baselines.
\end{enumerate}

%
\begin{figure*}[h]
\centering
\includegraphics[width=\textwidth]{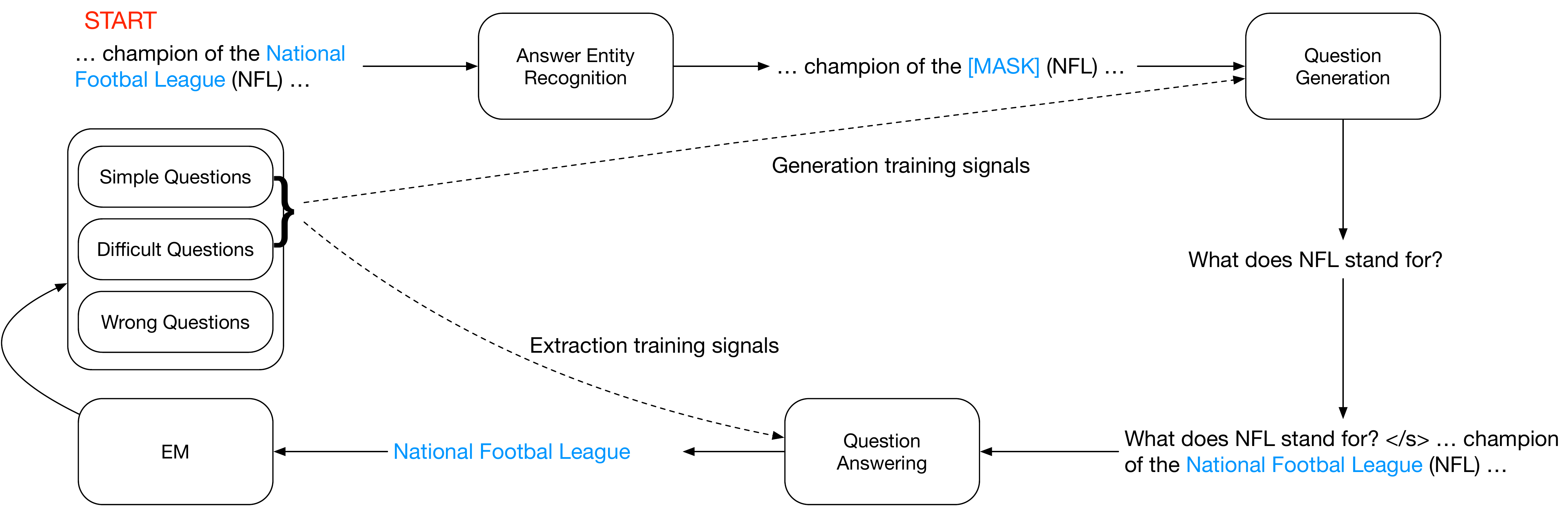}
\caption{The cooperative learning pipeline for question answering. The pipeline starts from a passage and follows the steps: $(1)$ recognizing a potential answer entity, $(2)$ generating a question asking about the answer entity, and $(3)$ answering the question by extracting the answer span in the passage.}
\label{fig:pipeline}
\end{figure*}
\section{Related Work}

Reinforcement learning and self-training have emerged recently for learning language generation in addition to maximum likelihood training. To optimize text generation models directly with non-differentiable objective functions, \citet{rennie2017self} proposed self-critical sequence training (SCST) using a policy gradient \cite{kakade2001natural,silver2014deterministic}. On the other hand, self-training has been shown to be effective in many tasks, such as machine translation \cite{he2019revisiting}, image classification \cite{xie2020self}, and structured database-grounded question answering \cite{xu2020autoqa}.

In the domain of question answering, a question generator can be used for joint answer prediction \cite{tang2017question,duan2017question}, and synthetic QA data are used for in-domain data augmentation \cite{sachan2018self,puri2020training,liu2020tell,klein2019learning} and out-of-domain adaptation. \citet{lewis2019unsupervised} and \citet{lee2020generating} introduced models for question answering under unsupervised/zero-shot settings. \citet{shakeri2020end} proposed generating synthetic question-answer pairs with an end-to-end model simultaneously. \citet{bartolo2021improving} improved the question synthesis by training with difficult QA cases from the AdversarialQA corpus \cite{bartolo2020beat} and fine-grained answer synthesis by multi-model voting. We include more related studies in Appendix \ref{sec:rw}.

In this work, we mainly compare our method with latest baselines, \citet{shakeri2020end} and \citet{bartolo2021improving} that reported results on out-of-domain adaptation. Besides improved QA performance, our framework, RGX, differs from the previous work in the following aspects: (1) Our method features reinforced finetuning of the QA Synthesizer, (2) Our framework supports and improves maximize mutual information inference in test time, and (3) Our work did not use complicated data annotation, e.g. AdversarialQA.

\section{RGX Framework}
In this section, we first introduce (1) the QA synthesis pipeline, (2) cooperative self-training for both QA synthesis and question answering, and (3) an improved maximum mutual information inference strategy. The self-training pipeline of RGX is shown in Figure \ref{fig:pipeline}.


\subsection{Data Synthesis}
Given a passage $p$, our goal is generating a set of questions $q$ and answers $a$ for the self-training of the QA model. The RGX model first recognize potential answer entities (AE) in $p$ with an answer entity recognition (AER) model, and then generate question based on the recognized AEs with a question generation (QG) model, and fine-grain the AEs with a pretrained question-answering extraction (QAE) model.



\subsubsection{Answer Entity Recognition (AER)}
\label{sec:aer-st}
Latest QA synthesis models, QAGen2S \cite{shakeri2020end} and SynQA \cite{bartolo2021improving}, directly generate questions from passages by modeling  $P_{qg}(q | p)$. In RGX, we first recognize all potential answer entities in a passage before generating questions for (1) increasing question diversity and coverage, and (2) modeling the mutual information between question generation and answering models in test time. The AER model in trained on the seed QA corpus.

We found that using an off-the-shelf named entity recognition (NER) model pretrained on the CONLL 2003 shared task~\cite{bender2003conll} performs poorly as a AER model (shown in our experiments). To learn an effective recognizer, given a passage $p$ and an annotated answer entity $e$, we select the sentence $s$ containing $e$ from $p$ and train language models to recognize $e$ in $s$. We tried two models for this task: a BIO sequence tagging model (AER-Tag) and a extractive AER model, which is similar to an extractive question answering model, for easier decoding. The model predicts the start and end positions of the answer entity $e$. With this method, we get potential answer entities by probabilities of all candidate spans. 

\subsubsection{Masked Question Generation}
With AER, we replace the answer entity $e$ in the passage $p$ with a [MASK] token and obtain the masked passage $p^*$. We then build a question generator $Q$ (denoted as QG interchangeably) that outputs answerable questions $q$ in natural language with the concatenation of $p^*$ and $e$ as input, i.e., $q = Q([p^*, e])$.
We adopt the BART sequence-to-sequence model \cite{lewis2019bart} as the architecture of $Q$ in our implementation, and we train $Q$ on the question-answer pairs in the seed corpus by maximizing the likelihood of annotated questions.

\subsubsection{Answer Extraction as Fine-grained AER}
The answer extraction model $A$ (denoted as QAE, question-answering extractor) takes generated question $q$ and the original passage $p$ as inputs. Following the standard extractive QA method, we predict the answers by
\begin{equation}
\label{eq:qa}
I_{st}, I_{ed} = A([q, p])
\end{equation}
where $I_{st}$ and $I_{ed}$ stand for the start and end positions of $e$ in $p$, respectively. We train the QAE model to predict $I_{st}$ and $I_{ed}$ separately with cross entropy losses.

Besides being trained with synthetic QA pairs and evaluated for the final QA performance, the QAE model is also a part of the data synthesis pipeline. After generating questions with the QG model, we use a pretrained QAE model to answer the generated questions. The QAE model recognizes better answers spans than the AER model since it takes questions as additional inputs. As a result, the final synthetic dataset is constructed by selecting generated questions and their corresponding QAE outputs. However, we still found the AER model necessary for generating diverse questions.


\subsection{Cooperative Self-training}
\label{sec:coop}
Although the pretrained models can generate synthetic QA pairs from corpora in unseen domains, there is always a domain shift from the seed QA corpus for pretraining to the target. To efficiently adapt the pretrained models to the new domains, we propose a cooperative self-training algorithm that allows finetuning on the target corpora without additional annotations. The finetuning is based on a three-agent (AER, QG, QAE) cooperative framework, RGX.
The pipeline is illustrated in Figure \ref{fig:pipeline} and comprises the following steps:
\begin{enumerate} \setlength{\itemsep}{0pt} \setlength{\parsep}{0pt}
\small
\item Produce a masked passage by replacing an answer entity selected by AER with the `[MASK]' token.
\item Generate a question asking about the masked entity.
\item Feed the generated question and the original passage into the QAE to predict an answer span.
\item Optimize the QAE model with selected QA pairs.
\item Optimize the QG model with selected QA pairs.
\end{enumerate}


In the proposed pipeline, all the AER, QG, and QAE models need pretraining to provide a reasonable start point for the cooperative self-training. However, the domain gap between the pretraining and the target corpus causes performance degradation. To mitigate the gap, we propose to measure the quality of generated questions and incorporate the measurement in loss functions. The quality is defined in two folds, correctness and difficulty.  Firstly, the question should be fluent and answerable, and secondly, it should not be too trivial.  
To automatically select high-quality generated QA pairs, we introduce a expectation-maximization (EM) method based on QAE losses that learns the question quality without supervision. 


\subsubsection{Synthetic QA Selection with EM}
To select synthetic QA pairs for finetuning, we first divide the generated questions based on the QAE loss for each question into three groups: low-, medium-, and high- loss questions. We can interpret questions with low loss as simple ones that the QAE model can easily answer. Medium-loss questions are challenging for the QAE, while those with high loss usually contain noise (e.g., containing grammatical errors or asking about incorrect answers). If we train the answering model with all questions, the training signal would be very noisy due to the high-loss questions. If we only reward questions that are correctly answered, the generator will converge to a trivial local optima. Thus, we train the QG and QAE model with the low- and medium- loss questions, namely simple and challenging questions. For the entire pipeline to be fully-automatic, we classify a given QA pair into one of the three types described above. Note that simply setting the thresholds as hyper-parameters is difficult since the loss decreases as the QAE model varies with different passages and domains. In order to find the thresholds adaptively, we apply an expectation-maximization (EM) algorithm to bucket synthetic QA pairs for each passage.

We finetune both QG and QAE models with the selected simple and challenging QA pairs. After the training, re-running the RGX pipeline with the finetuned question generation model leads to improved data synthesis. Training the QAE model on the updated synthetic dataset can significant outperform the previous finetuned QAE model.

\subsubsection{Maximum Mutual Information QA}
\citet{li2016mutual} proposed a maximum mutual information (MMI) decoding method for machine translation, and \citet{tang2017question} proposed a MMI method for jointly learning question generation and answering models. There is no previous study to our knowledge that applies MMI inference in test time of question answering that improves the final performance, because (1) modeling $P(q|p, a)$ for all possible answers (spans) $a$ is too inefficient, and (2) Unlike the QAE model that receives loss signals from all words in a given passage, the QG model does not receive loss signal from the passage directly, so $P_{qg}(q|p, a)$ it is less accurate for ranking answer spans.

However, the AER and self-training strategy enable efficient MMI inference for QA,
\[
a = \underset{a}{\mathrm{argmax}} [\alpha \* \log P_{qg}(q | p, a) + \beta \* \log P_{qa} (a | p, q)]
\]
In test time, we run the RGX pipeline for each passage without additional training to get fine-grained AEs and corresponding questions. On the other hand, we take the top span predicted by the QAE model, and the top-k answer entities spans recognized by the RGX pipeline. In practice, we fix $\beta = 1$. We used an adaptive $\alpha$ value by comparing the synthetic question generated by the QG model and the input question. For each answer entity $a$, we calculate
\[
\alpha = \texttt{max} (1 - \texttt{abs} (\frac{q_{input}}{q_{gen}} - 1), 0.1)
\]
This value normalizes the question probability $p(q|p, a)$ estimated by the QG model, since generated questions from some answer entities is easier than other spans in the same passage, which makes the QG model assign all natural questions a relative low perplexity.


\section{Experiments}
In this work, we train three modules for building the cooperative self-training environment RGX, i.e., the answer entity recognizer (AER), the question generator (QG), and the question-answering extractor (QAE). We used a BERT \cite{devlin2018bert} model for AER, a BART \cite{lewis2019bart} model for QG, and an ELECTRA \cite{clark2019electra} model for AER and QAE. To compare with the results reported in \citet{shakeri2020end} and \citet{bartolo2021improving}, we (1) pretrain question generation and answering models on the seed corpora, (2) generate synthetic QA data on the target domains, (3) finetune QA models with synthetic data, and (4) evaluate the finetuned QA model on human-labeled evaluation sets. The source code and demo are publicly available\footnote{\url{https://github.com/luohongyin/RGX}}.


\subsection{Data}
In our experiment work, we leveraged Natural Questions \cite{kwiatkowski2019natural} and SQuAD v1.1 \cite{rajpurkar2016SQuAD} as the seed corpora for pretraining all modules introduced above. To evaluate the performance of the proposed RGX on question answering tasks with different difficulty levels, we conduct experiments on both SQuAD v1.1 \cite{rajpurkar2016SQuAD} and MRQA \cite{fisch2019mrqa} out-of-domains (BioASQ, TextbookQA, RACE, RelationExtraction, DuoRC, and DROP). In the following sections, we use the term SQuAD to represent the SQuAD v1.1 corpus. For self-training, we sample 3000 passages from the training set of each corpus for data synthesis. More details about the data are provided in Appendix \ref{sec:ddetail}


\subsection{Implementation Details}

\noindent \textbf{Pretraining.}
We pretrain the AER, QG, and QAE models on NaturalQuestions and SQuAD (i.e., the seed) corpora. For NaturalQuestions, we only use the data points containing a short answer. For Cooperative training, we follow the steps described in Section \ref{sec:coop} for the cooperative training phase. 


\noindent \textbf{Self-training.}
We apply self-training for QG and QAE by finetuning the models on selected synthetic QA pairs using the same method as pretraining. The AER model is fixed after pretraining. The QAE model is finetuned using the official Huggingface \cite{wolf2019huggingface} training scripts for question answering. We will open-source the RGX framework if the submission is accepted. 

\noindent \textbf{Hyperparameters.} There are three phases of model training in this work: pretraining on the seed corpora, cooperative adaptation with self-training on the target corpora, and final finetuning on the synthetic data. We adopt most of the hyper-parameters reported in the original BERT \cite{devlin2018bert}, BART \cite{lewis2019bart}, and ELECTRA \cite{clark2019electra} papers.
We select the final finetuning learning rates from $\left \{ 3e-5, 4e-5, 5e-5 \right \}$ and report the highest performance. All the other hyper-parameters are the same as reported in the corresponding papers. For all the phases, we fix $eps=1e-6$ and $s_{w}=2000$, where $s_w$ is the number of warm-up steps, and we apply no weight decays. We use BART-large (406M parameters) and ELECTRA-large (335M parameters) models for our experiments. We run our experiments on 2 Tesla V100 GPUs. Training the QAE models on augmented data takes about 4 hours.

\subsection{Experiment Results}
We assess the performance of RGX with both semi-annotated and zero-annotated evaluation on unseen domains using exact match (EM) and F1 scores. The exact match metric assesses the percentage of predicted spans that are exactly the same as labeled answers, while the F1 score measure the overall token-level overlap between predicted and labeled answers. In our semi-annotated setting, we use the annotated answer entities in the target corpora but utilize QG to generate questions for obtaining the training question-answer pairs. The labeled questions are not used. We employ no annotation from the target corpora for the out-of-domain task but automatically construct the question-answer training pairs with entities and questions inferred by AER and QG on the corpora.


\subsubsection{Semi-annotated Evaluation}
The model performance with the pretrained QA model, RGX, and SOTA trained with full-supervision is shown in Table \ref{tab:semi}.

%
\begin{table}[h]
\small
\centering
\begin{tabular}{@{}lll@{}}
\toprule
\textbf{Models}             & \textbf{EM}   & \textbf{F1}   \\ \midrule
\multicolumn{3}{c}{Source domain: NQ, Target domain: SQuAD}      \\ \hdashline[1.5pt/2pt]
ELECTRA-large (NaturalQuestions)    & 67.8 & 80.3 \\
RGX  & 83.1 & 90.7 \\
$\quad$--w/o Coop. ST    & 81.2 & 89.1 \\
ELECTRA-large (SQuAD)     & 89.7 & 94.9 \\ \bottomrule
\end{tabular}
\caption{The performance of the question answering models in the semi-annotated setting. RGX stands for our cooperative training approach, and Coop. ST stands for cooperative self-training.}
\label{tab:semi}
\end{table}

\begin{table*}[t]
\small
\centering
\begin{tabular}{@{}lllllllllllllll@{}}
\toprule
\multicolumn{1}{c}{$\underset{Domain}{\textbf{Model}}$} & \multicolumn{2}{c}{$\underset{Bio}{\textbf{BioASQ}}$}                      & \multicolumn{2}{c}{$\underset{Book}{\textbf{TextbookQA}}$}                  & \multicolumn{2}{c}{$\underset{Exam}{\textbf{RACE}}$}                        & \multicolumn{2}{c}{$\underset{Wiki}{\textbf{RelExt.}}$}                          & \multicolumn{2}{c}{$\underset{Movie}{\textbf{DuoRC}}$}                       & \multicolumn{2}{c}{$\underset{Wiki}{\textbf{DROP}}$}   & \multicolumn{2}{c}{\textbf{Avg}}                      \\ \midrule
\multicolumn{1}{c}{}      & \multicolumn{1}{c}{\small EM} & \multicolumn{1}{c}{\small F1} & \multicolumn{1}{c}{\small EM} & \multicolumn{1}{c}{\small F1} & \multicolumn{1}{c}{\small EM} & \multicolumn{1}{c}{\small F1} & \multicolumn{1}{c}{\small EM} & \multicolumn{1}{c}{\small F1} & \multicolumn{1}{c}{\small EM} & \multicolumn{1}{c}{\small F1} & \multicolumn{1}{c}{\small EM} & \multicolumn{1}{c}{\small F1} & \multicolumn{1}{c}{\small EM} & \multicolumn{1}{c}{\small F1} \\ \hdashline[1.5pt/2pt]
\multicolumn{15}{c}{Source Domain: NaturalQuestions$_{wiki}$, Method: Extraction}                                                                                                                                                                                                                                                                                 \\ \hdashline[1.5pt/2pt]
{\small ELECTRA}             & 41.9                   & 59.0                   & 31.9                   & 41.5                   & 32.4                   & 43.4                   & 67.7                   & 81.8                   & 40.0                   & 48.5                   & \textbf{39.3}                   & \textbf{51.1} & 42.2 & 54.2                  \\
{\small QAGen2S}       & 43.2                   & 64.1                   & 39.9                   & 51.7                   & 33.7                   & 45.5                   & 71.6                   & 84.4                   & 43.8                   & 53.2                   & 24.2                   & 37.1 & 42.7 & 56.0                    \\
{\small RGX (Ours)}       & \textbf{50.3}                   & \textbf{70.1}                   & \textbf{49.9}                   & \textbf{60.9}                   & \textbf{40.3}                   & \textbf{52.4}                   & \textbf{76.1}                   & \textbf{87.2}                   & \textbf{47.8}                   & \textbf{58.4}                   & 27.6                   & 42.1 & \textbf{48.7} & \textbf{61.9}                   \\
{\small $\quad$-- w/o MMI}             & 49.7          & 69.1          & 49.4          & 60.6          & 39.7          & 51.5          & 75.4          & 86.7          & 46.9          & 57.5          & 27.1          & 41.7 & 46.8 & 61.2                       \\
{\small $\quad$-- w/o EM}                     & 48.2                   & 67.9                   & 47.4                   & 59.8                   & 38.3                   & 50.5                   & 74.1                   & 86.2                   & 46.6                   & 56.9                   & 26.1                   & 40.9 & 46.8 & 60.4          \\
{\small $\quad$-- w/o CST}                     & 45.4                   & 66.4                   & 41.9                   & 53.8                   & 35.1                   & 47.2                   & 72.7                   & 85.4                   & 45.5                   & 54.9                   & 24.6                   & 37.9 & 44.2 & 57.6         \\ \midrule
\multicolumn{15}{c}{Source Domain: SQuAD$_{wiki}$ (SQuAD+AQA+Wiki for SynQA), Method: Extraction}                                                                                                                                                                                                                                                                                             \\ \hdashline[1.5pt/2pt]
{\small ELECTRA}        & 58.7                   & 73.1                   & 43.0                   & 53.6                   & 38.3                   & 52.5                   & 79.0                   & 88.4                   & 53.1                   & 64.2                   & 48.3                   & 60.8 & 53.4 & 65.4                   \\
{\small QAGen2S}       & 56.8                   & 71.7                   & 48.0                   & 56.5                   & 43.4                   & 54.9                   & 73.4                   & 84.8                   & 53.3                   & 64.6                   & 42.2                   & 54.5 & 52.8 & 64.5                  \\
{\small SynQA}       & 55.1                   & 68.7                   & 41.4                   & 50.2                   & 40.2                   & 54.2                   & 78.9                   & \textbf{88.6}                   & 51.7                   & 62.1                   & \textbf{64.9}                   & \textbf{73.0} & 55.3 & 66.1                  \\
{\small RGX (Ours)}       & \textbf{60.3}                   & \textbf{74.8}                   & \textbf{51.2}                   & \textbf{61.2}                   & \textbf{44.9}                   & \textbf{58.7}                   & \textbf{79.2}                   & \textbf{88.6}                   & \textbf{57.4}                   & \textbf{66.2}                   & 47.6                   & 60.9 & \textbf{56.8} & \textbf{68.4}                  \\
{\small $\quad$-- w/o MMI}                     & 59.2          & 73.6          & 50.1          & 60.4          & 46.3          & 57.6          & 78.9          & 88.5          & 56.2                       & 65.7                       & 46.9                       & 60.6 & 56.3 & 67.7                      \\
{\small $\quad$-- w/o EM}             & 52.1                       & 64.0                       & 50.6                       & 58.9                       & 35.4                       & 48.3                       & 75.6                       & 85.9                       & 55.6                       & 64.9                       & 40.7                       & 53.2 & 51.7 & 62.5                       \\
{\small $\quad$-- w/o CST}             & 57.5                   & 72.1                   & 48.6                   & 57.0                   & 43.8                   & 55.2                   & 74.3                   & 85.3                   & 53.9                   & 65.3                   & 43.0                   & 55.1 & 53.5 & 65.0                  \\
\midrule
\multicolumn{15}{c}{Source Domain: SQuAD$_{wiki}$, Method: Prompt Tuning + Seq2seq Generation} \\ \hdashline[1.5pt/2pt]
{\small T5}             & 54.6                       & 71.1                       & 37.9                       & 61.9                       & 15.0                       & 53.1                       & 74.5                       & 86.5                       & 48.2                       & 65.2                       & 40.4                       & 51.9 & 45.1 & 64.9                       \\
{\small T5 + RGX}             & 55.1                       & 71.6                       & 41.1                       & 64.2                       & 15.5                       & 55.1                       & 75.9                       & 87.1                       & 49.5                       & 66.2                       & 42.9                       & 53.8 & 46.7 & 66.3                      \\
\bottomrule
\end{tabular}
\caption{The QA performance evaluation on the out-of-domains of the MRQA benchmark. All models used are pretrained on the human-labeled training set from the source domains, and the QA models are finetuned on synthetic data generated based on the unannotated passages of the target domains. The finetuned QA models are evaluated on human-generated evaluation data for each target domains with the exact match (EM) and F1 scores. MMI stands for maximum mutual information inference, EM stands for involving difficult questions with EM selection, and CST stands for cooperative self-training.}
\label{tab:mrqa}
\end{table*}

\begin{table}[]
\small
\begin{tabular}{@{}llll@{}}
\toprule
                   & \textbf{QAGen2S}   & \textbf{SynQA}         & \textbf{RGX}           \\ \midrule
Pretraining & XQ     & SQ+AQA        & XQ         \\
Synthesis   & Target    & Wikipedia     & Target        \\
Finetuning    & XQ+Syn    & SQ+AQA+Syn     & XQ+Syn        \\
AER Model          & None      & None          & ELECTRA       \\
Coop. ST & No        & No            & Yes           \\
QA Num.       & 1M        & 1.5M          & 0.3M          \\ \bottomrule
\end{tabular}
\caption{Comparison of different self-training methods. XQ stands for ``NaturalQuestions (NQ) or SQuAD (SQ)''. QA Num. stands for the number of synthetic QA pairs used for self-training.}
\label{tab:comp}
\end{table}

Table \ref{tab:semi} shows that RGX yields improvement over the pretrained model, approaching the SOTA performance of the fully trained ELECTRA-large-discriminator model. The experiment result suggests that the cooperative learning strategy improves the question generation model with human-annotated answer entities.


\subsubsection{Out-of-domain Evaluation}
We also evaluate the models in unseen domains, where we do not use any annotated QA for finetuning. We train the QAE models based on the synthetic training data and evaluate the models on the target domains. We compare RGX with latest self-training QA methods, QAGen2S \cite{shakeri2020end} and SynQA \cite{bartolo2021improving}. Since QAGen2S did not report full MRQA results, we implemented our own version. We first present the RGX performance and the results reported by the authors QAGen2S and SynQA, and then conduct ablation study by training different language models on RGX synthetic QA data.


The full evaluation results on MRQA out-of-domains are shown in Table \ref{tab:mrqa}, and the experiment setting comparison is shown in table \ref{tab:comp}. The results show that the models trained with the RGX framework achieve significantly higher EM and F1 scores on most domains, comparing to both pretrained QA models and self-training baselines. The results showed that the RGX model achieves 7.7 and 3.0 average F1 improvement over ELECTRA, the SOTA pretrained language model for QA, by pretraining on NQ and SQuAD respectively. The improvement over previous SOTA self-training QA methods, QAGen2S and SynQA, is also significant on both pretraining corpora, although SynQA applies complicated adversarial QA annotation. The largest gain we got is adapting NQ model to TextbookQA domain, increasing 18.0 EM and 19.4 F1 scores. Note that our model still outperforms all baselines without MMI. The performance on the DROP benchmark drops since DROP requires multi-step reasoning, but the synthetic generation model tends to generate safe question-answer pairs. We also found that without selecting harder questions with SEM in RGX, the performance is significantly lower. These facts indicate that the QA model needs hard training examples for better performance, and explains the good performance of SynQA on DROP. For the same reason, the performance drop led by removing EM from RGX is significantly larger when the QG model is pretrained on SQuAD, since SQuAD questions are more coherent with the context than NQ, and selecting simple questions for RGX training will encourage the model to generate trivial questions, which is harmful for the QA training.


\begin{table}[t]
\small
\centering
\begin{tabular}{@{}lll@{}}
\toprule
\textbf{Models}             & \textbf{EM}   & \textbf{F1}   \\ \midrule
\multicolumn{3}{c}{Source domain: NQ, Target domain: SQuAD}      \\ \hdashline[1.5pt/2pt]
Pretrained NQ & 67.8 & 80.3 \\
RGX + NER  & 27.4 & 35.4 \\
RGX + AER-Tag  & 71.4 & 82.4 \\
RGX + AER-LM  & 72.7 & 85.9 \\
RGX + AER-EM  & 79.2 & 89.4 \\
Supervised ELECTRA-large      & 89.7 & 94.9 \\ \bottomrule
\end{tabular}
\caption{Comparison of different AER strategies. NER stands for the BERT named entity recognition model trained on the CONLL 2003 shared task. AER-Tag stands for a BIO-based tagging strategy, AER-LM means selecting synthetic QA pairs with lowest QAE losses. AER-EM is the EM-based QA selection strategy applied in our full model.}
\label{tab:nq-sq}
\end{table}

\begin{table}[]
\small
\begin{tabular}{@{}lllllll@{}}
\toprule
           & \multicolumn{2}{c}{\textbf{ELECTRA}}            & \multicolumn{2}{c}{\textbf{Top-k+MMI}}          & \multicolumn{2}{c}{\textbf{AER+MMI}}            \\ \midrule
           & \multicolumn{1}{c}{EM} & \multicolumn{1}{c}{F1} & \multicolumn{1}{c}{EM} & \multicolumn{1}{c}{F1} & \multicolumn{1}{c}{EM} & \multicolumn{1}{c}{F1} \\
           \hdashline[1.5pt/2pt]
BioASQ     & 58.7                   & 73.1                   & 57.8                       & 72.9                       & 59.9                   & 74.0                   \\
TextbookQA & 43.0                   & 54.6                   & 44.6                       & 54.9                       & 45.3                   & 55.4                   \\
RACE       & 38.3                   & 52.5                   & 38.1                       & 52.4                       & 39.7                       & 54.1                       \\
RelExt     & 79.0                   & 88.4                   & 78.6                       & 88.3                       & 79.2                       & 88.6                       \\
DuoRC      & 53.1                   & 64.2                   &  52.6                      & 64.3                       & 53.8                       & 65.1                       \\
DROP       & 48.3                   & 60.8                   & 46.7                       & 60.8                       & 49.7                       & 61.5                       \\ \bottomrule
\end{tabular}
\caption{Comparison between maximum mutual information inference performance grounded on AER results and top-k ($k=20$) predictions of the QA model.}
\label{tab:aer-mmi}
\end{table}

\begin{table}[h]
\small
\centering
\begin{tabular}{@{}llll@{}}
\toprule
\multicolumn{1}{c}{\textbf{Models}} & \multicolumn{1}{c}{\textbf{Mean Len.}} & \multicolumn{1}{c}{\textbf{Std Len.}} & \multicolumn{1}{c}{\textbf{Vocab}} \\ \midrule

Ground-truth               & 11.29                          & 3.72                          & 988703                         \\
Semi-anno. RGX               & 10.54                          & 1.91                          & 923191                         \\
$\quad$--w/o Coop. ST               & 10.49                          & 2.48                          & 919105                         \\
Zero-anno. RGX               & 10.53                          & 1.94                          & 873300                         \\
$\quad$--w/o Coop. ST               & 10.57                          & 2.63                          & 789924                         \\
$\quad$--w/o AER               & 10.60                          & 1.87                          & 743454                        \\
$\quad$--w/o EM               & 10.18                          & 1.62                          & 692301                          \\
\bottomrule
\end{tabular}
\caption{The vocabulary sizes and lengths of Annotated and generated questions on SQuAD under both semi- and zero-annotated settings in unseen domains}
\label{tab:ques-stat}
\end{table}

\begin{figure}[t]
\centering
\includegraphics[width=\columnwidth]{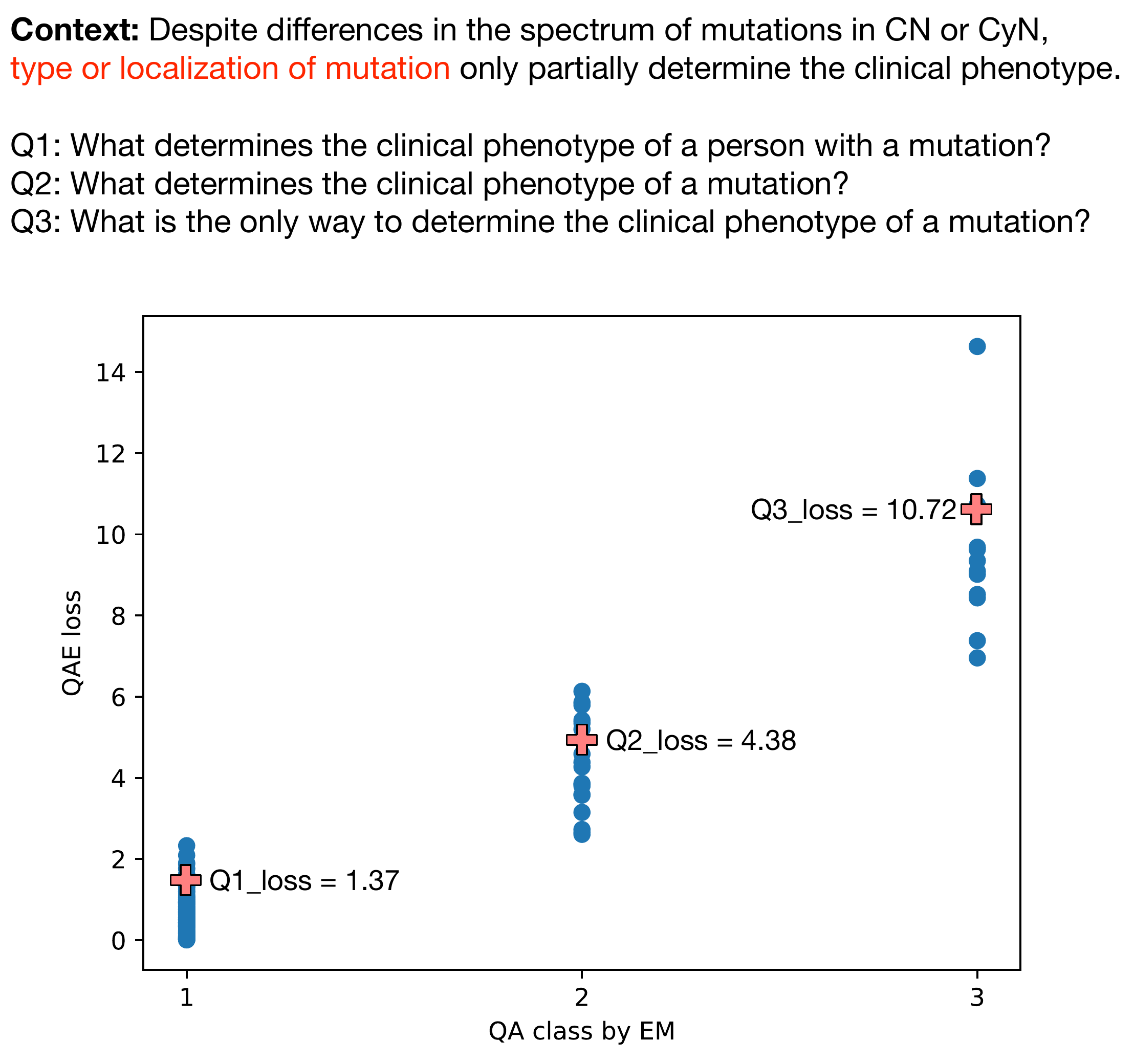}
\caption{Generated questions about the same answer entity classified into different types by EM. Questions Q1 is answered by the QAE model confidently, while the Q2 is considered more challenging than Q1 since less information is provided. Q3 is an unanswerable questions given the context passage.}
\label{fig:em_example}
\end{figure}

\begin{table}[]
\small
\centering
\begin{tabular}{@{}lllll@{}}
\toprule
\multicolumn{1}{c}{\textbf{Domain}} & \multicolumn{2}{c}{\textbf{RGX w/o Coop. ST}} & \multicolumn{2}{c}{\textbf{RGX}} \\ \midrule
                           & Hit        & BLEU      & Hit         & BLEU        \\ \hdashline[1.5pt/2pt]
BioASQ                     & 68.1       & 5.9      & 75.8         & 12.7        \\
TextbookQA                 & 43.7       & 7.5      & 58.2         & 13.2        \\
RACE                       & 8.3        & 5.2      & 12.3         & 6.8         \\
RelExt.                         & 47.4        & 2.8       & 54.2         & 3.3         \\
DuoRC                      & 53.5       & 6.7          & 60.0         & 7.5            \\
DROP                       & 73.5       & 12.3          & 75.3        & 9.3            \\ \bottomrule
\end{tabular}
\caption{Evaluation of the answer hit rates and question BLEU scores of the synthetic data. Hit rate stands for the percentage of human-labeled answer entities in the evaluation passages that are successfully covered by the selected synthetic data generated by RGX.}
\label{tab:syn-eval}
\end{table}



\begin{table*}[t]
\small
\begin{tabular}{@{}lcc@{}}
\toprule
\multicolumn{3}{l}{\begin{tabular}[c]{@{}l@{}}Architecturally, the school has a Catholic character. Atop \textcolor{orange}{the Main Building}'s gold dome is \textcolor{orange}{a golden statue of the Virgin}\\ \textcolor{orange}{Mary}. Immediately in front of the Main Building and facing it, is \textcolor{orange}{a copper statue of Christ} with arms upraised with the\\ legend "\textcolor{blue}{Venite Ad Me Omnes}". Next to the Main Building is the Basilica of the Sacred Heart. \textcolor{red}{Immediately behind the}\\ \textcolor{red}{basilica is the Grotto}, \textcolor{orange}{a Marian place of prayer and reflection}.  It is \textcolor{red}{a replica} of \textcolor{blue}{the grotto at Lourdes, France} where the\\ Virgin Mary reputedly appeared to \textcolor{orange}{Saint Bernadette Soubirous} in \textcolor{blue}{1858}. At the end of the main drive (and in a direct line\\ that connects through 3 statues and the Gold Dome), is \textcolor{blue}{a simple, modern stone statue of Mary}.\end{tabular}} \\ \midrule
\multicolumn{1}{c}{\textbf{Annotated}}                                                                                                                                                                                                         & \textbf{Pretrained}                                                                                                                                                                                                                                                            & \textbf{RGX}                                                                                                                                                                                                                                             \\ \hdashline[1.5pt/2pt]
\multicolumn{1}{c}{\textbf{Saint Bernadette Soubirous}}                                                                                                                                                                                          & \textbf{\begin{tabular}[c]{@{}c@{}}a Marian place of\\ prayer and reflection\end{tabular}}                                                                                                                                                                                     & \textbf{\begin{tabular}[c]{@{}c@{}}a Marian place of\\ prayer and reflection\end{tabular}}                                                                                                                                                                 \\
\begin{tabular}[c]{@{}l@{}}To whom did the Virgin Mary allegedly\\ appear in 1858 in Lourdes France?\end{tabular}                                                                                                                                & \multicolumn{1}{l}{what is the grotto at st bernadette's?}                                                                                                                                                                                                                     & \multicolumn{1}{l}{\begin{tabular}[c]{@{}l@{}}what is the grotto in st bernadette\\ school?\end{tabular}}                                                                                                                                                  \\
\multicolumn{1}{c}{\textbf{a copper statue of Christ}}                                                                                                                                                                                           & \textbf{\begin{tabular}[c]{@{}c@{}}the grotto at Lourdes,\\ France\end{tabular}}                                                                                                                                                                                               & \textbf{Venite Ad Me Omnes}                                                                                                                                                                                                                                \\
\begin{tabular}[c]{@{}l@{}}What is in front of the Notre Dame\\ Main Building?\end{tabular}                                                                                                                                                      & \multicolumn{1}{l}{\begin{tabular}[c]{@{}l@{}}where is the grotto located at st\\ bernadette school?\end{tabular}}                                                                                                                                                             & \multicolumn{1}{l}{\begin{tabular}[c]{@{}l@{}}what is the message on the statue in\\ front of st bernadette school?\end{tabular}}                                                                                                                          \\
\multicolumn{1}{c}{\textbf{the Main Building}}                                                                                                                                                                                                   & \textbf{\begin{tabular}[c]{@{}c@{}}Immediately behind the\\ basilica is the Grotto\end{tabular}}                                                                                                                                                                               & \textbf{1858}                                                                                                                                                                                                                                              \\
\begin{tabular}[c]{@{}l@{}}The Basilica of the Sacred heart at\\ Notre Dame is beside to which structure?\end{tabular}                                                                                                                           & \multicolumn{1}{l}{what is the grotto in st peter's school?}                                                                                                                                                                                                                   & \multicolumn{1}{l}{when was the grotto at lourdes built?}                                                                                                                                                                                                  \\
\multicolumn{1}{c}{\textbf{\begin{tabular}[c]{@{}c@{}}a Marian place of\\ prayer and reflection\end{tabular}}}                                                                                                                                   & \textbf{\begin{tabular}[c]{@{}c@{}}copper statue of Christ\\ with arms upraised\end{tabular}}                                                                                                                                                                                  & \textbf{\begin{tabular}[c]{@{}c@{}}a simple, modern\\ stone statue of Mary\end{tabular}}                                                                                                                                                                   \\
What is the Grotto at Notre Dame?                                                                                                                                                                                                                & \multicolumn{1}{l}{what is it a statue of christ?}                                                                                                                                                                                                                             & \multicolumn{1}{l}{\begin{tabular}[c]{@{}l@{}}what is the statue at st bernadette\\ school?\end{tabular}}                                                                                                                                                  \\
\multicolumn{1}{c}{\textbf{\begin{tabular}[c]{@{}c@{}}a golden statue of\\ the Virgin Mary\end{tabular}}}                                                                                                                                        & \textbf{a replica}                                                                                                                                                                                                                                                             & \textbf{\begin{tabular}[c]{@{}c@{}}the grotto at Lourdes,\\ France\end{tabular}}                                                                                                                                                                           \\
\begin{tabular}[c]{@{}l@{}}What sits on top of the Main\\ Building at Notre Dame?\end{tabular}                                                                                                                                                   & \multicolumn{1}{l}{\begin{tabular}[c]{@{}l@{}}is the grotto at st bernadette school\\ in paris a replica of which European\\ landmark?\end{tabular}}                                                                                                                           & \multicolumn{1}{l}{\begin{tabular}[c]{@{}l@{}}what is the replica of st bernadette's\\ school in paris?\end{tabular}}                                                                                                                                      \\ \bottomrule
\end{tabular}
\caption{An example of a passage in the training set of the SQuAD corpus. We list the annotated question-answer pairs, and the question-answer pairs generated by the models pretrained on NQ and finetuned by RGX. The bold texts are annotated or recognized answer entities. Adapting from NQ is difficult since the questions in NQ do not strictly coherent with a given context. More generation examples are shown in Appendix \ref{sec:examples}.}
\label{tab:case-study}
\end{table*}

\subsection{Analysis}

\subsubsection{Answer Entity Recognition}
We first compare the performance of different AER models and strategies by setting NQ as the source domain and SQuAD 1.1 as the target domain in Table \ref{tab:nq-sq}. The results showed that the choice of AER model and strategy significantly influences the final QA performance. The low performance of the NER model trained on CONLL shared task suggests the importance of our AER module. We notice that the improvement from the cooperative learning over the pretrained models is higher in the zero-annotation setting than the semi-annotated task. The observation indicates that the model trained with RGX is more robust against the automatically recognized answer entities. More details about the AER methods are shown in Appendix \ref{sec:aer}.

The AER method also enables and improves the maximum mutual information (MMI) inference in test time. Table \ref{tab:mrqa} shows that MMI achieves the best performance, and we also show that the MMI accuracy is hurt without AER. Table \ref{tab:aer-mmi} shows that MMI grounded on AER constantly outperform the ELECTRA model, but grounding on top-k seriously hurts the EM scores. Some invalid answer predictions leads to low question generation perplexities, which makes MMI inference noisy. Table \ref{tab:ques-stat} shows that the QG model generated more diverse questions based on the AER outputs.

\subsubsection{Synthetic QA Selection with EM}
Previous experiments showed that selecting non-trivial synthetic QA pairs is essential for RGX to achieve high performance. Table \ref{tab:mrqa} shows that the performance of cooperative self-trained RGX is much lower than the pretrained baseline without EM. If selecting QA pairs with low perplexities instead of EM, the QA diversity is significantly lower as shown in Table \ref{tab:ques-stat}, thus makes the QAE model overfit to simple training cases and hurts the QA accuracy. We show questions about the same answer entity being classified into simple, challenging, and difficult types by EM in figure \ref{fig:em_example}. The data points in the plot represents the losses of synthetic QA pairs and the predicted QA type. Based on the highlighted answer entity, question 1 and 2 are predicted as correct questions, while question 3, which has a relatively high QAE loss, is regarded as a wrong question. Note that we only generate one question for each span recognized by the AER model, but different questions might be re-directed to the same AE after QAE fine-graining.

\subsubsection{Cooperative Self-training}
We found that the cooperative self-training method improves domain adaptation ability of self-trained QA models by increasing both accuracy and diversity of QA synthesis.

\noindent \textbf{Accuracy.} We also evaluate the quality of the generated QA pairs without a downstream task by assessing the answer entity hit rate and the BLEU scores of generated questions using the evaluation sets of each domain. The results are shown in Table \ref{tab:syn-eval}, indicating that RGX find mores human-annotated answer entities, and the generated questions have higher BLEU scores on all domains. The evaluation results show that the synthetic QA pars generated by RGX covers more human annotated answer entities, and the generated questions are more similar to human annotations than the pretrained question generation model. We also found that tuning the generation model for more than 1 iterations does not result in further improvement, since keeping training language models with their own outputs leads to difficult optimization.

\noindent \textbf{Diversity.}
We compare the lengths and vocabulary sizes of the questions and summarize the statistics in Table \ref{tab:ques-stat}, which shows that the ground-truth questions are longer and more diverse in vocabulary than the generated ones. However, the cooperative self-training, together with AER and EM, improves the vocabulary diversity.
We observe a correlation between the vocabulary size and the QA performance reported in Table \ref{tab:semi} and \ref{tab:nq-sq}, presumably because the QAE model requires diverse knowledge for training. Thus, we believe generating more diverse QA pairs with good quality will be a critical next step to improve RGX.

\noindent \textbf{Case Study.} An example of a SQuAD passage is shown in Table \ref{tab:case-study}. We list the annotated and generated question-answer pairs by different models. The table shows that the models can recognize reasonable answer entities other than the annotated ones, and RGX generates more natural QAs.


\section{Conclusion}
We propose a cooperative self-training framework, RGX, consisting of an answer entity Recognizer, a question Generator, and an answer eXtractor, for question generation and answering. We also introduce in the framework an expectation-maximization method that measures the quality of generated questions for reinforced finetuning of the question generation models. Experiments show that RGX significantly outperforms pretrained and self-trained model baselines while adapted to unseen domains, suggesting that RGX is a promising framework for making extractive question answering methods more scalable and less dependent on human annotation.


\bibliographystyle{acl_natbib}
\bibliography{anthology,acl2021}

\begin{thebibliography}{48}
\expandafter\ifx\csname natexlab\endcsname\relax\def\natexlab#1{#1}\fi

\bibitem[{Bartolo et~al.(2020)Bartolo, Roberts, Welbl, Riedel, and
  Stenetorp}]{bartolo2020beat}
Max Bartolo, Alastair Roberts, Johannes Welbl, Sebastian Riedel, and Pontus
  Stenetorp. 2020.
\newblock Beat the ai: Investigating adversarial human annotation for reading
  comprehension.
\newblock \emph{Transactions of the Association for Computational Linguistics},
  8:662--678.

\bibitem[{Bartolo et~al.(2021)Bartolo, Thrush, Jia, Riedel, Stenetorp, and
  Kiela}]{bartolo2021improving}
Max Bartolo, Tristan Thrush, Robin Jia, Sebastian Riedel, Pontus Stenetorp, and
  Douwe Kiela. 2021.
\newblock Improving question answering model robustness with synthetic
  adversarial data generation.
\newblock \emph{arXiv preprint arXiv:2104.08678}.

\bibitem[{Bender et~al.(2003)Bender, Och, and Ney}]{bender2003conll}
Oliver Bender, Franz~Josef Och, and Hermann Ney. 2003.
\newblock Maximum entropy models for named entity recognition.
\newblock In \emph{Proceedings of CoNLL-2003}, pages 148--151. Edmonton,
  Canada.

\bibitem[{Bengio et~al.(2003)Bengio, Ducharme, Vincent, and
  Jauvin}]{bengio2003neural}
Yoshua Bengio, R{\'e}jean Ducharme, Pascal Vincent, and Christian Jauvin. 2003.
\newblock A neural probabilistic language model.
\newblock \emph{Journal of machine learning research}, 3(Feb):1137--1155.

\bibitem[{Clark et~al.(2019)Clark, Luong, Le, and Manning}]{clark2019electra}
Kevin Clark, Minh-Thang Luong, Quoc~V Le, and Christopher~D Manning. 2019.
\newblock Electra: Pre-training text encoders as discriminators rather than
  generators.
\newblock In \emph{International Conference on Learning Representations}.

\bibitem[{Devlin et~al.(2018)Devlin, Chang, Lee, and
  Toutanova}]{devlin2018bert}
Jacob Devlin, Ming-Wei Chang, Kenton Lee, and Kristina Toutanova. 2018.
\newblock Bert: Pre-training of deep bidirectional transformers for language
  understanding.
\newblock \emph{arXiv preprint arXiv:1810.04805}.

\bibitem[{Duan et~al.(2017)Duan, Tang, Chen, and Zhou}]{duan2017question}
Nan Duan, Duyu Tang, Peng Chen, and Ming Zhou. 2017.
\newblock Question generation for question answering.
\newblock In \emph{Proceedings of the 2017 Conference on Empirical Methods in
  Natural Language Processing}, pages 866--874.

\bibitem[{Fei et~al.(2020)Fei, Ren, and Ji}]{fei2020retrofitting}
Hao Fei, Yafeng Ren, and Donghong Ji. 2020.
\newblock Retrofitting structure-aware transformer language model for end
  tasks.
\newblock \emph{arXiv preprint arXiv:2009.07408}.

\bibitem[{Fisch et~al.(2019)Fisch, Talmor, Jia, Seo, Choi, and
  Chen}]{fisch2019mrqa}
Adam Fisch, Alon Talmor, Robin Jia, Minjoon Seo, Eunsol Choi, and Danqi Chen.
  2019.
\newblock Mrqa 2019 shared task: Evaluating generalization in reading
  comprehension.
\newblock In \emph{Proceedings of the 2nd Workshop on Machine Reading for
  Question Answering}, pages 1--13.

\bibitem[{Glass et~al.(2019)Glass, Gliozzo, Chakravarti, Ferritto, Pan,
  Bhargav, Garg, and Sil}]{glass2019span}
Michael Glass, Alfio Gliozzo, Rishav Chakravarti, Anthony Ferritto, Lin Pan,
  GP~Bhargav, Dinesh Garg, and Avirup Sil. 2019.
\newblock Span selection pre-training for question answering.
\newblock \emph{arXiv preprint arXiv:1909.04120}.

\bibitem[{Goodfellow(2016)}]{goodfellow2016nips}
Ian Goodfellow. 2016.
\newblock Nips 2016 tutorial: Generative adversarial networks.
\newblock \emph{arXiv preprint arXiv:1701.00160}.

\bibitem[{Goodfellow et~al.(2014)Goodfellow, Pouget-Abadie, Mirza, Xu,
  Warde-Farley, Ozair, Courville, and Bengio}]{goodfellow2014generative}
Ian Goodfellow, Jean Pouget-Abadie, Mehdi Mirza, Bing Xu, David Warde-Farley,
  Sherjil Ozair, Aaron Courville, and Yoshua Bengio. 2014.
\newblock Generative adversarial nets.
\newblock \emph{Advances in neural information processing systems},
  27:2672--2680.

\bibitem[{Havrylov and Titov(2017)}]{havrylov2017emergence}
Serhii Havrylov and Ivan Titov. 2017.
\newblock Emergence of language with multi-agent games: Learning to communicate
  with sequences of symbols.
\newblock In \emph{Advances in neural information processing systems}, pages
  2149--2159.

\bibitem[{He et~al.(2019)He, Gu, Shen, and Ranzato}]{he2019revisiting}
Junxian He, Jiatao Gu, Jiajun Shen, and Marc'Aurelio Ranzato. 2019.
\newblock Revisiting self-training for neural sequence generation.
\newblock \emph{arXiv preprint arXiv:1909.13788}.

\bibitem[{Henderson et~al.(2019)Henderson, Casanueva, Mrk{\v{s}}i{\'c}, Su,
  Wen, and Vuli{\'c}}]{henderson2019convert}
Matthew Henderson, I{\~n}igo Casanueva, Nikola Mrk{\v{s}}i{\'c}, Pei-Hao Su,
  Tsung-Hsien Wen, and Ivan Vuli{\'c}. 2019.
\newblock Convert: Efficient and accurate conversational representations from
  transformers.
\newblock \emph{arXiv preprint arXiv:1911.03688}.

\bibitem[{Humeau et~al.(2019)Humeau, Shuster, Lachaux, and
  Weston}]{humeau2019poly}
Samuel Humeau, Kurt Shuster, Marie-Anne Lachaux, and Jason Weston. 2019.
\newblock Poly-encoders: Transformer architectures and pre-training strategies
  for fast and accurate multi-sentence scoring.
\newblock \emph{arXiv preprint arXiv:1905.01969}.

\bibitem[{Jia et~al.(2021)Jia, Lewis, and Zettlemoyer}]{jia2021question}
Robin Jia, Mike Lewis, and Luke Zettlemoyer. 2021.
\newblock Question answering infused pre-training of general-purpose
  contextualized representations.
\newblock \emph{arXiv preprint arXiv:2106.08190}.

\bibitem[{Joshi et~al.(2020)Joshi, Chen, Liu, Weld, Zettlemoyer, and
  Levy}]{joshi2020spanbert}
Mandar Joshi, Danqi Chen, Yinhan Liu, Daniel~S Weld, Luke Zettlemoyer, and Omer
  Levy. 2020.
\newblock Spanbert: Improving pre-training by representing and predicting
  spans.
\newblock \emph{Transactions of the Association for Computational Linguistics},
  8:64--77.

\bibitem[{Kakade(2001)}]{kakade2001natural}
Sham~M Kakade. 2001.
\newblock A natural policy gradient.
\newblock \emph{Advances in neural information processing systems},
  14:1531--1538.

\bibitem[{Klein and Nabi(2019)}]{klein2019learning}
Tassilo Klein and Moin Nabi. 2019.
\newblock Learning to answer by learning to ask: Getting the best of gpt-2 and
  bert worlds.
\newblock \emph{arXiv preprint arXiv:1911.02365}.

\bibitem[{Kwiatkowski et~al.(2019)Kwiatkowski, Palomaki, Redfield, Collins,
  Parikh, Alberti, Epstein, Polosukhin, Devlin, Lee
  et~al.}]{kwiatkowski2019natural}
Tom Kwiatkowski, Jennimaria Palomaki, Olivia Redfield, Michael Collins, Ankur
  Parikh, Chris Alberti, Danielle Epstein, Illia Polosukhin, Jacob Devlin,
  Kenton Lee, et~al. 2019.
\newblock Natural questions: a benchmark for question answering research.
\newblock \emph{Transactions of the Association for Computational Linguistics},
  7:453--466.

\bibitem[{Lazaridou et~al.(2016)Lazaridou, Peysakhovich, and
  Baroni}]{lazaridou2016multi}
Angeliki Lazaridou, Alexander Peysakhovich, and Marco Baroni. 2016.
\newblock Multi-agent cooperation and the emergence of (natural) language.
\newblock \emph{arXiv preprint arXiv:1612.07182}.

\bibitem[{Lee et~al.(2020)Lee, Lee, Jeong, Kim, and Hwang}]{lee2020generating}
Dong~Bok Lee, Seanie Lee, Woo~Tae Jeong, Donghwan Kim, and Sung~Ju Hwang. 2020.
\newblock Generating diverse and consistent qa pairs from contexts with
  information-maximizing hierarchical conditional vaes.
\newblock \emph{arXiv preprint arXiv:2005.13837}.

\bibitem[{Lewis et~al.(2019{\natexlab{a}})Lewis, Liu, Goyal, Ghazvininejad,
  Mohamed, Levy, Stoyanov, and Zettlemoyer}]{lewis2019bart}
Mike Lewis, Yinhan Liu, Naman Goyal, Marjan Ghazvininejad, Abdelrahman Mohamed,
  Omer Levy, Ves Stoyanov, and Luke Zettlemoyer. 2019{\natexlab{a}}.
\newblock Bart: Denoising sequence-to-sequence pre-training for natural
  language generation, translation, and comprehension.
\newblock \emph{arXiv preprint arXiv:1910.13461}.

\bibitem[{Lewis et~al.(2019{\natexlab{b}})Lewis, Denoyer, and
  Riedel}]{lewis2019unsupervised}
Patrick Lewis, Ludovic Denoyer, and Sebastian Riedel. 2019{\natexlab{b}}.
\newblock Unsupervised question answering by cloze translation.
\newblock \emph{arXiv preprint arXiv:1906.04980}.

\bibitem[{Lewis et~al.(2021)Lewis, Wu, Liu, Minervini, K{\"u}ttler, Piktus,
  Stenetorp, and Riedel}]{lewis2021paq}
Patrick Lewis, Yuxiang Wu, Linqing Liu, Pasquale Minervini, Heinrich
  K{\"u}ttler, Aleksandra Piktus, Pontus Stenetorp, and Sebastian Riedel. 2021.
\newblock Paq: 65 million probably-asked questions and what you can do with
  them.
\newblock \emph{arXiv preprint arXiv:2102.07033}.

\bibitem[{Li and Jurafsky(2016)}]{li2016mutual}
Jiwei Li and Dan Jurafsky. 2016.
\newblock Mutual information and diverse decoding improve neural machine
  translation.
\newblock \emph{arXiv preprint arXiv:1601.00372}.

\bibitem[{Liu et~al.(2020)Liu, Gong, Fu, Yan, Chen, Lv, Duan, and
  Zhou}]{liu2020tell}
Dayiheng Liu, Yeyun Gong, Jie Fu, Yu~Yan, Jiusheng Chen, Jiancheng Lv, Nan
  Duan, and Ming Zhou. 2020.
\newblock Tell me how to ask again: Question data augmentation with
  controllable rewriting in continuous space.
\newblock \emph{arXiv preprint arXiv:2010.01475}.

\bibitem[{Liu et~al.(2019)Liu, Ott, Goyal, Du, Joshi, Chen, Levy, Lewis,
  Zettlemoyer, and Stoyanov}]{liu2019roberta}
Yinhan Liu, Myle Ott, Naman Goyal, Jingfei Du, Mandar Joshi, Danqi Chen, Omer
  Levy, Mike Lewis, Luke Zettlemoyer, and Veselin Stoyanov. 2019.
\newblock Roberta: A robustly optimized bert pretraining approach.
\newblock \emph{arXiv preprint arXiv:1907.11692}.

\bibitem[{Luo et~al.(2019)Luo, Jiang, Belinkov, and Glass}]{luo2019improving}
Hongyin Luo, Lan Jiang, Yonatan Belinkov, and James Glass. 2019.
\newblock Improving neural language models by segmenting, attending, and
  predicting the future.
\newblock \emph{arXiv preprint arXiv:1906.01702}.

\bibitem[{Mikolov et~al.(2013)Mikolov, Sutskever, Chen, Corrado, and
  Dean}]{mikolov2013distributed}
Tomas Mikolov, Ilya Sutskever, Kai Chen, Greg~S Corrado, and Jeff Dean. 2013.
\newblock Distributed representations of words and phrases and their
  compositionality.
\newblock \emph{Advances in neural information processing systems},
  26:3111--3119.

\bibitem[{Mordatch and Abbeel(2018)}]{mordatch2018emergence}
Igor Mordatch and Pieter Abbeel. 2018.
\newblock Emergence of grounded compositional language in multi-agent
  populations.
\newblock In \emph{Proceedings of the AAAI Conference on Artificial
  Intelligence}, volume~32.

\bibitem[{Pennington et~al.(2014)Pennington, Socher, and
  Manning}]{pennington2014glove}
Jeffrey Pennington, Richard Socher, and Christopher~D Manning. 2014.
\newblock Glove: Global vectors for word representation.
\newblock In \emph{Proceedings of the 2014 conference on empirical methods in
  natural language processing (EMNLP)}, pages 1532--1543.

\bibitem[{Peters et~al.(2018)Peters, Neumann, Iyyer, Gardner, Clark, Lee, and
  Zettlemoyer}]{peters2018deep}
Matthew~E Peters, Mark Neumann, Mohit Iyyer, Matt Gardner, Christopher Clark,
  Kenton Lee, and Luke Zettlemoyer. 2018.
\newblock Deep contextualized word representations.
\newblock \emph{arXiv preprint arXiv:1802.05365}.

\bibitem[{Puri et~al.(2020)Puri, Spring, Patwary, Shoeybi, and
  Catanzaro}]{puri2020training}
Raul Puri, Ryan Spring, Mostofa Patwary, Mohammad Shoeybi, and Bryan Catanzaro.
  2020.
\newblock Training question answering models from synthetic data.
\newblock \emph{arXiv preprint arXiv:2002.09599}.

\bibitem[{Rajpurkar et~al.(2016)Rajpurkar, Zhang, Lopyrev, and
  Liang}]{rajpurkar2016SQuAD}
Pranav Rajpurkar, Jian Zhang, Konstantin Lopyrev, and Percy Liang. 2016.
\newblock Squad: 100,000+ questions for machine comprehension of text.
\newblock \emph{arXiv preprint arXiv:1606.05250}.

\bibitem[{Rennie et~al.(2017)Rennie, Marcheret, Mroueh, Ross, and
  Goel}]{rennie2017self}
Steven~J Rennie, Etienne Marcheret, Youssef Mroueh, Jerret Ross, and Vaibhava
  Goel. 2017.
\newblock Self-critical sequence training for image captioning.
\newblock In \emph{Proceedings of the IEEE Conference on Computer Vision and
  Pattern Recognition}, pages 7008--7024.

\bibitem[{Sachan and Xing(2018)}]{sachan2018self}
Mrinmaya Sachan and Eric Xing. 2018.
\newblock Self-training for jointly learning to ask and answer questions.
\newblock In \emph{Proceedings of the 2018 Conference of the North American
  Chapter of the Association for Computational Linguistics: Human Language
  Technologies, Volume 1 (Long Papers)}, pages 629--640.

\bibitem[{Shakeri et~al.(2020)Shakeri, Santos, Zhu, Ng, Nan, Wang, Nallapati,
  and Xiang}]{shakeri2020end}
Siamak Shakeri, Cicero Nogueira~dos Santos, Henry Zhu, Patrick Ng, Feng Nan,
  Zhiguo Wang, Ramesh Nallapati, and Bing Xiang. 2020.
\newblock End-to-end synthetic data generation for domain adaptation of
  question answering systems.
\newblock \emph{arXiv preprint arXiv:2010.06028}.

\bibitem[{Shen et~al.(2018)Shen, Lin, Jacob, Sordoni, Courville, and
  Bengio}]{shen2018straight}
Yikang Shen, Zhouhan Lin, Athul~Paul Jacob, Alessandro Sordoni, Aaron
  Courville, and Yoshua Bengio. 2018.
\newblock Straight to the tree: Constituency parsing with neural syntactic
  distance.
\newblock \emph{arXiv preprint arXiv:1806.04168}.

\bibitem[{Shen et~al.(2020)Shen, Tay, Zheng, Bahri, Metzler, and
  Courville}]{shen2020structformer}
Yikang Shen, Yi~Tay, Che Zheng, Dara Bahri, Donald Metzler, and Aaron
  Courville. 2020.
\newblock Structformer: Joint unsupervised induction of dependency and
  constituency structure from masked language modeling.
\newblock \emph{arXiv preprint arXiv:2012.00857}.

\bibitem[{Silver et~al.(2014)Silver, Lever, Heess, Degris, Wierstra, and
  Riedmiller}]{silver2014deterministic}
David Silver, Guy Lever, Nicolas Heess, Thomas Degris, Daan Wierstra, and
  Martin Riedmiller. 2014.
\newblock Deterministic policy gradient algorithms.

\bibitem[{Tang et~al.(2017)Tang, Duan, Qin, Yan, and Zhou}]{tang2017question}
Duyu Tang, Nan Duan, Tao Qin, Zhao Yan, and Ming Zhou. 2017.
\newblock Question answering and question generation as dual tasks.
\newblock \emph{arXiv preprint arXiv:1706.02027}.

\bibitem[{Vaswani et~al.(2017)Vaswani, Shazeer, Parmar, Uszkoreit, Jones,
  Gomez, Kaiser, and Polosukhin}]{vaswani2017attention}
Ashish Vaswani, Noam Shazeer, Niki Parmar, Jakob Uszkoreit, Llion Jones,
  Aidan~N Gomez, {\L}ukasz Kaiser, and Illia Polosukhin. 2017.
\newblock Attention is all you need.
\newblock In \emph{Advances in neural information processing systems}, pages
  5998--6008.

\bibitem[{Wolf et~al.(2019)Wolf, Debut, Sanh, Chaumond, Delangue, Moi, Cistac,
  Rault, Louf, Funtowicz et~al.}]{wolf2019huggingface}
Thomas Wolf, Lysandre Debut, Victor Sanh, Julien Chaumond, Clement Delangue,
  Anthony Moi, Pierric Cistac, Tim Rault, R{\'e}mi Louf, Morgan Funtowicz,
  et~al. 2019.
\newblock Huggingface's transformers: State-of-the-art natural language
  processing.
\newblock \emph{arXiv preprint arXiv:1910.03771}.

\bibitem[{Xie et~al.(2020)Xie, Luong, Hovy, and Le}]{xie2020self}
Qizhe Xie, Minh-Thang Luong, Eduard Hovy, and Quoc~V Le. 2020.
\newblock Self-training with noisy student improves imagenet classification.
\newblock In \emph{Proceedings of the IEEE/CVF Conference on Computer Vision
  and Pattern Recognition}, pages 10687--10698.

\bibitem[{Xu et~al.(2020)Xu, Semnani, Campagna, and Lam}]{xu2020autoqa}
Silei Xu, Sina~J Semnani, Giovanni Campagna, and Monica~S Lam. 2020.
\newblock Autoqa: From databases to qa semantic parsers with only synthetic
  training data.
\newblock \emph{arXiv preprint arXiv:2010.04806}.

\bibitem[{Yu et~al.(2017)Yu, Zhang, Wang, and Yu}]{yu2017seqgan}
Lantao Yu, Weinan Zhang, Jun Wang, and Yong Yu. 2017.
\newblock Seqgan: Sequence generative adversarial nets with policy gradient.
\newblock In \emph{Proceedings of the AAAI conference on artificial
  intelligence}, volume~31.

\end{thebibliography}

\newpage
\appendix

\section{More Related Work}
\label{sec:rw}
Representation learning has been an important topic in NLP area since neural language models were proposed \cite{bengio2003neural}. Based on word co-occurrence, \citet{mikolov2013distributed} and \citet{pennington2014glove} proposed language embedding algorithms to model word-level semantics. Recent studies have focused on pretraining contextualized word representations with large-scaled corpora \cite{peters2018deep}. State-of-the-art representation models are pretrained with the masked language modeling task \cite{devlin2018bert,liu2019roberta,clark2019electra} using the Transformer architecture \cite{vaswani2017attention}.

Different variants of masked language models have been investigated to improve performance in downstream tasks. \citet{joshi2020spanbert} leveraged a masked span generation task instead of word prediction. \citet{fei2020retrofitting} and \citet{shen2020structformer} proposed models that learns better syntax knowledge with syntactic distances \cite{shen2018straight} and heights \cite{luo2019improving}. \citet{henderson2019convert} and \citet{humeau2019poly} showed that pretraining language models on dialog corpora perform better on dialog-related downstream tasks, as compared to pretraining on Wikipedia. A span selection pretraining objective is proposed in \citet{glass2019span} to reduce the gap between the pretraining and downstream finetuning stages and to improve the performance on the QA task. Some applications of generated questions are shown in \cite{lewis2021paq,jia2021question}.

In contrast to self-training methods that usually adopt a teacher-student learning strategy, cooperative learning pipelines contain several agents working together to learn as much knowledge as possible. A typical cooperative learning framework is generative adversarial networks (GAN) \cite{goodfellow2016nips,goodfellow2014generative}, where a generator is optimized to confuse a discriminator, and a discriminator is trained to distinguish real examples from generated ones. Sequence GAN is further designed for learning diverse text generation \cite{yu2017seqgan}. Unlike the adversarial learning method where two networks work for opposite goals, other studies proposed learning environments in which different agents learn the same objective functions for language emergence \cite{lazaridou2016multi,mordatch2018emergence,havrylov2017emergence}, including simple natural language, compositional language, and symbolic language.

\section{Data}
\label{sec:ddetail}
The SQuAD v1.1 is the easiest QA corpus used in this paper. The dataset contains $107,785$ question-answer pairs on $536$ articles, which are split into passages. Each question is labeled with an answer that can be extracted from the given passage.

The Natural Questions dataset is a large-scale corpus designed for open-domain question answering. The dataset is more challenging than SQuAD. All questions are collected from human search queries and are annotated with long and abstractive answers. Some of the questions are also labeled with a short answer for learning answer-span extraction or reading comprehension. Focusing on the machine reading comprehension task, we select $106,926$ questions labeled with both long and short answers from the dataset for experiments.

For each target domain in MRQA, we collect the corresponding training data and sample 3000 passages for QA synthesis. The number of synthetic QAs varies based on the length of input passages, and is shown in Table \ref{tab:mrqa-num}.

\begin{table}[]
\centering
\begin{tabular}{@{}lc@{}}
\toprule
\textbf{Dataset} & \multicolumn{1}{l}{\textbf{Num. Synthetic QA}} \\ \midrule
BioASQ           & 123121                                         \\
TextbookQA       & 133773                                         \\
RACE             & 115847                                         \\
RelExt.          & 52142                                          \\
DuoRC            & 250698                                         \\
DROP             & 100394                                         \\ \bottomrule
\end{tabular}
\caption{Number of synthetic QA of each MRQA domain.}
\label{tab:mrqa-num}
\end{table}

\section{Answer Entity Recognition Details}
\label{sec:aer}
In this section, we describe details of the AER methods, which are not covered in detail in previous sections. All AER models are pretrained on the Natural Questions corpus. To solve the sparsity problem, in other words, the passages are long but not all potential question-answer pairs are annotated, we train all following AER models by using the sentence containing the annotated answer entities as inputs, instead of the whole passage. If a sentence in the passage does not contain an annotated answer entity, we do not use it for training.

In this work, we introduce two types of AER methods, tagging based AER (AER-tag) and extraction based AER (AER-Search and AER-Coop). We describe their training and how we use the trained model to recognize answer entities in our experiments.

\subsection{AER-Tag}

\subsubsection{Training}
We apply a BIO tagging model for answer entity recognition in the AER-Tab method. We train the model to classify all tokens in the input sentence into three classes,
\begin{itemize}
\item B(egin) - the first token of the annotated answer entity
\item I(nsize) - other tokens of the annotated answer entity
\item O(utside) - tokens that are not a part of the annotated answer entity
\end{itemize}

\subsubsection{Evaluation}
Given an input passage, we run the trained BIO tagging model on each of its sentences and greedily predict answer entities. There might be more than one answer entities predicted in each sentence, and we only use the answer entities start with a predicted B tag.

\subsection{AER-LM}

\subsubsection{Training}
For AER-LM method, we need to pretrain an extraction-based AER model. We also take a sentence of $L$ tokens containing an annotated answer entity as an example. Using an extraction model, which is similar as our question answering model, we train the model to predict the start and end location of the annotated answer entity. The model outputs a start score and an end score for each token, and predicts the start/end locations by selecting the tokens that are assigned with highest scores. The model is trained with cross-entropy loss, by regarding the extraction task as two $L$-class classification tasks.

\subsubsection{Evaluation}
In evaluation, we first run the model on each sentence of the input passages and calculate the start and end scores for each token. For each span $(x_i, x_{i+1}, \dots, x_j)$ that is not longer than $L_{span}$ tokens, we calculate the span score with
\begin{equation}
s_{ij} = s_{st}^i + s_{ed}^j
\end{equation}
where $s_{st}^i$ is the start score of the first token of span $(i, j)$, and $s_{ed}^j$ is the end score of the last token of the span. In practice, we set $L_{span} = 10$.

To re-rank all possible answer entities, we select top $N_0 = 40$ spans according to $s_{ij}$ for each passage. For all selected answer entities, we generated questions with a pretrained question generator and collect the generation perplexity of the questions. We select $N_{search} = 5$ question-answer pairs with lowest perplexities for the final question-answering finetuning.

\subsection{AER-Coop}
In AER-Coop, we use the same extraction training method applied in AER-Search, and we also use the $s_{ij}$ scores to select the top $N_0 = 40$ preliminary answer entities for further search. The difference is that we search for final answer entities cooperatively with the pretrained question generator and question answering extractor.

With the question generator and question answering extractor, we re-rank the recognized answer entities with the following score
\begin{equation}
s_{ij}^c = \gamma \cdot I_c - p
\end{equation}
where $\gamma$ is a large, positive coefficient, $p$ is the perplexity of generated question based on span $(i, j)$, and $I_c = 1$ if the generated question is correctly answered, and otherwise $I_c = 0$.

\subsection{Answer Entity Overlapping}
We found the extraction-based AER model leads to overlapping problems, since a large start or end score assigned to a token leads to many candidate answer entities start or end at the token. In practice, if an answer entity is selected by the AER-Search and AER-Coop method, we no longer consider any other answer entities that overlap with the selected ones.

\section{RGX Examples}
\label{sec:examples}
In this section, we show some examples of our full model. The examples are contained in Table \ref{tab:gen-examples}.

\begin{table*}[t]
\small
\begin{tabular}{@{}lll@{}}
\toprule
\multicolumn{3}{l}{\begin{tabular}[c]{@{}l@{}}The National History Museum of Montevideo is located in the historical residence of General Fructuoso Rivera. It exhi-\\ bits artifacts related to the history of Uruguay. In a process begun in 1998, the National Museum of Natural History (1837)\\ and the National Museum of Anthropology (1981), merged \textbf{\textcolor{blue}{in 2001}}, becoming the National Museum of Natural History\\ and Anthropology. In July 2009, the two institutions again became independent. The Historical Museum has annexed eight\\ historical houses in the city, five of which are located in the Ciudad Vieja. One of them, on the same block with the main\\ building, is the historic residence of Antonio Montero, which houses the Museo Romantico.\end{tabular}}                                                                                                                                                                                                                                                                                             \\\hdashline[1.5pt/2pt] 
\multicolumn{3}{l}{\textbf{When was the national history museum of montevideo founded?}}                                                                                                                                                                                                                                                                                                                                                                                                                                                                                                                                                                                                                                                                                                                                                                                                                                                                                                                                                                                         \\\midrule
\multicolumn{3}{l}{\begin{tabular}[c]{@{}l@{}}In the 1920s, John Maynard Keynes prompted a division between microeconomics and macroeconomics. Under Keynesian\\ economics macroeconomic trends can overwhelm economic choices made by individuals. Governments should promote\\ aggregate demand for goods as a means to encourage economic expansion. Following World War II, Milton Friedman\\ created the concept of monetarism. Monetarism focuses on using the \textbf{\textcolor{blue}{supply and demand of money}} as a method for con-\\ trolling economic activity. In the 1970s, monetarism has adapted into supply-side economics which advocates reducing\\ taxes as a means to increase the amount of money available for economic expansion.\end{tabular}}                                                                                                                                                                                                                                                                                                                                   \\\hdashline[1.5pt/2pt]
\multicolumn{3}{l}{\textbf{Monarism focuses on the relationship between the?}}                                                                                                                                                                                                                                                                                                                                                                                                                                                                                                                                                                                                                                                                                                                                                                                                                                                                                                                                                                               \\\midrule
\multicolumn{3}{l}{\begin{tabular}[c]{@{}l@{}}Starting in 2006, Apple's industrial design shifted to favor aluminum, which was used in the construction of the first Mac-\\ Book Pro. Glass was added in 2008 with the introduction of the unibody MacBook Pro. These materials are billed as env-\\ ironmentally friendly. The iMac, MacBook Pro, MacBook Air, and Mac Mini lines currently all use aluminum enclosures,\\ and are now made of a single unibody. Chief designer \textbf{\textcolor{blue}{Jonathan Ive}} continues to guide products towards a minimalist and\\ simple feel, including eliminating of replaceable batteries in notebooks. Multi-touch gestures from the iPhone's interface\\ have been applied to the Mac line in the form of touch pads on notebooks and the Magic Mouse and Magic Trackpad for\\ desktops.\end{tabular}}                                                                                                                                                                                                                                                  \\\hdashline[1.5pt/2pt]
\multicolumn{3}{l}{\textbf{Who is the designer of the macbook pro?}}                                                                                                                                                                                                                                                                                                                                                                                                                                                                                                                                                                                                                                                                                                                                                                                                                                                                                                                                                                                                       \\\midrule
\multicolumn{3}{l}{\begin{tabular}[c]{@{}l@{}}The city's total area is 468.9 square miles (1,214 km2). 164.1 sq mi (425 km2) of this is water and 304.8 sq mi (789 km2) is\\ land. The highest point in the city is Todt Hill \textbf{\textcolor{blue}{on Staten Island}}, which, at 409.8 feet (124.9 m) above sea level, is the\\ highest point on the Eastern Seaboard south of Maine. The summit of the ridge is mostly covered in woodlands as part\\ of the Staten Island Greenbelt.\end{tabular}}                                                                                                                                                                                                                                                                                                                                                                                                                                                                                                                                                                                                    \\\hdashline[1.5pt/2pt]
\multicolumn{3}{l}{\textbf{Where is the highest point in new york city?}}                                                                                                                                                                                                                                                                                                                                                                                                                                                                                                                                                                                                                                                                                                                                                                                                                                                                                                                                                                                              \\\midrule
\multicolumn{3}{l}{\begin{tabular}[c]{@{}l@{}}In 1922, the number of supporters had surpassed 20,000 and by lending money to the club, Barça was able to build the\\ larger Camp de Les Corts, which had an initial capacity of 20,000 spectators. After the Spanish Civil War the club started\\ attracting more members and a larger number of spectators at matches. This led to several expansion projects: the\\ grandstand in 1944, the southern stand in 1946, and finally the northern stand in 1950. After the last expansion, Les Corts\\ could hold \textbf{\textcolor{blue}{60,000 spectators}}.\end{tabular}}                                                                                                                                                                                                                                                                                                                                                                                                                                                                                  \\\hdashline[1.5pt/2pt]
\multicolumn{3}{l}{\textbf{What is the capacity of barcelona's stadium?}}                                                                                                                                                                                                                                                                                                                                                                                                                                                                                                                                                                                                                                                                                                                                                                                                                                                                                                                                                                                             \\\midrule
\multicolumn{3}{l}{\begin{tabular}[c]{@{}l@{}}On 1 November 2013, international postal services for Somalia officially resumed. \textbf{\textcolor{blue}{The Universal Postal Union}} is now\\ assisting the Somali Postal Service to develop its capacity, including providing technical assistance and basic mail\\ processing equipment.\end{tabular}}                                                                                                                                                                                                                                                                                                                                                                                                                                                                                                                                                                                                                                                                                                                                                   \\\hdashline[1.5pt/2pt]
\multicolumn{3}{l}{\textbf{Who is responsible for supporting the somali postal service?}}                                                                                                                                                                                                                                                                                                                                                                                                                                                                                                                                                                                                                                                                                                                                                                                                                                                                                                                                                                    \\\midrule
\multicolumn{3}{l}{\begin{tabular}[c]{@{}l@{}}In addition to membership, as of 2010{[}update{]} there are 1,335 officially registered fan clubs, called penyes, around the\\ world. The fan clubs promote Barcelona in their locality and receive beneficial offers when visiting Barcelona. Among\\ the best supported teams globally, Barcelona has the highest social media following in the world among sports teams,\\ with over 90 million Facebook fans as of February 2016. The club has had many prominent people among its support-\\ ers, including \textbf{\textcolor{blue}{Pope John Paul II}}, who was an honorary member, and former prime minister of Spain José Luis\\ Rodríguez Zapatero. FC Barcelona has the second highest average attendance of European football clubs only behind\\ Borussia Dortmund.\end{tabular}}                                                                                                                                                                                                                                                                \\\hdashline[1.5pt/2pt]
\multicolumn{3}{l}{\textbf{Who was an honorary member of barcelona football club?}}                                                                                                                                                                                                                                                                                                                                                                                                                                                                                                                                                                                                                                                                                                                                                                                                                                                                                                                                                                                    \\\midrule
\multicolumn{3}{l}{\begin{tabular}[c]{@{}l@{}}In April 1758, the British concluded the Anglo-Prussian Convention with Frederick in which they committed to pay him\\ \textbf{\textcolor{blue}{an annual subsidy of £670,000}}. Britain also dispatched 9,000 troops to reinforce Ferdinand's Hanoverian army, the first\\ British troop commitment on the continent and a reversal in the policy of Pitt. Ferdinand had succeeded in driving the\\ French from Hanover and Westphalia and re-captured the port of Emden in March 1758 before crossing the Rhine with\\ his own forces, which caused alarm in France. Despite Ferdinand's victory over the French at the Battle of Krefeld and\\ the brief occupation of Düsseldorf, he was compelled by the successful manoeuvering of larger French forces to with-\\ draw across the Rhine.\end{tabular}}                                                                                                                                                                                                                                                 \\\hdashline[1.5pt/2pt]
\multicolumn{3}{l}{\textbf{What did france pay to the prussian monarchy?}}                                                                                                                                                                                                                                                                                                                                                                                                                                                                                                                                                                                                                                                                                                                                                                                                                                                                                                                                                                                 \\\midrule
\multicolumn{3}{l}{\begin{tabular}[c]{@{}l@{}}Executives at Trump Entertainment Resorts, whose sole remaining property will be the Trump Taj Mahal, said in 2013\\ that they were considering the option of selling the Taj and \textbf{\textcolor{blue}{winding down and exiting the gaming and hotel business}}.\end{tabular}}                                                                                                                                                                                                                                                                                                                                                                                                                                                                                                                                                                                                                                                                                                                                                                            \\\hdashline[1.5pt/2pt]
\multicolumn{3}{l}{\textbf{What is the future of the trump taj mahal?}}                                                                                                                                                                                                                                                                                                                                                                                                                                                                                                                                                                                                                                                                                                                                                                                                                                                                                                                                                           \\\midrule
\multicolumn{3}{l}{\begin{tabular}[c]{@{}l@{}}Vehicles typically include headlamps and tail lights. Headlamps are white or selective yellow lights placed in the front of\\ the vehicle, designed to illuminate the upcoming road and to make the vehicle more visible. Many manufactures are turn-\\ ing to LED headlights as an energy-efficient alternative to traditional headlamps. Tail and brake lights are red and emit\\ light to the rear so as to reveal the vehicle's direction of travel to following drivers. White rear-facing reversing lamps in-\\ dicate that the vehicle's transmission has been placed in the reverse gear, warning anyone behind the vehicle that it is\\ moving backwards, or about to do so. Flashing turn signals on the front, side, and rear of the vehicle indicate an intended\\ change of position or direction. In \textbf{\textcolor{blue}{the late 1950s}}, some automakers began to use electroluminescent technology to back-\\ light their cars' speedometers and other gauges or to draw attention to logos or other decorative elements.\end{tabular}} \\\hdashline[1.5pt/2pt]
\multicolumn{3}{l}{\textbf{When did they start putting back up lights in cars?}}                                                                                                                                                                                                                                                                                                                                                                                                                                                                                                                                                                                                                                                                                                                                                                                                                                                                                                                                                                                         \\ \bottomrule
\end{tabular}
\caption{Examples of recognized answer entities and generated questions with the full RGX model}
\label{tab:gen-examples}
\end{table*}

\end{document}